\newif\ifconfver
\confverfalse     

\newif\ifplainver  
\plainvertrue

\ifplainver
    \confverfalse   
\fi

\ifconfver
\newcommand{\CLASSINPUTtoptextmargin}{2.54cm}
\newcommand{\CLASSINPUTbottomtextmargin}{1.91cm}

\newcommand{\CLASSINPUTinnersidemargin}{1.91cm}
\newcommand{\CLASSINPUToutersidemargin}{1.91cm}
\documentclass[conference]{IEEEtran}
\else
    \ifplainver
        \documentclass[11pt]{article}
        \usepackage{fullpage}
    \else
        \documentclass[11pt,draftcls,onecolumn]{IEEEtran}
    \fi
\fi
\usepackage{calc,amsfonts,amssymb,amsmath,bm,url,color,theorem,graphicx,cite,shortcuts_FW}
\usepackage{psfrag,subfigure,float,hyperref}
\usepackage{algorithm}
\usepackage{algorithmic}
\usepackage{multirow}

\usepackage[font=small,labelfont=bf,tableposition=top]{caption}

\definecolor{orange}{RGB}{255,107,0}

\newtheorem{Lemma}{Lemma}

\newtheorem{Theorem}{Theorem}
\newtheorem{Assumption}{Assumption}

\usepackage{pgfplots}
\pgfplotsset{compat=1.3}
\usepackage{tikz}
\usetikzlibrary{arrows,shapes,calc,tikzmark,backgrounds,matrix,decorations.markings}
\usepgfplotslibrary{fillbetween}

\usetikzlibrary{shadows.blur}
\definecolor{asuorange}{rgb}{1,0.699,0.0625}
\definecolor{asured}{rgb}{0.598,0,0.199}
\definecolor{asuborder}{rgb}{0.953,0.484,0}
\definecolor{asugrey}{rgb}{0.309,0.332,0.340}
\definecolor{asublue}{rgb}{0,0.555,0.836}
\definecolor{asugold}{rgb}{1,0.777,0.008}

\definecolor{lavander}{cmyk}{0,0.48,0,0}
\definecolor{violet}{cmyk}{0.79,0.88,0,0}
\definecolor{burntorange}{cmyk}{0,0.52,1,0}

\tikzstyle{stubborn}=[draw,circle, black!80, fill=black!40,
                       text=violet,minimum width=10pt]
\tikzstyle{superpeers}=[draw,circle, asublue!80!white, fill = asublue!50!white,
                       text=violet,minimum width=10pt]
\tikzstyle{susceptible}=[draw,circle, left color = red, color = red,
                       text=violet,minimum width=10pt]
\tikzstyle{perturb}=[draw,circle,burntorange, left color=blue,
                       text=violet,minimum width=10pt]
\tikzstyle{legend_general}=[rectangle, rounded corners, thin,
                           black, fill= white, draw, text=black,
                           minimum width=2.5cm, minimum height=0.8cm]
\tikzstyle{legend_graph}=[rectangle, rounded corners, thin,
                           black, fill= white, draw, text=black,
                           minimum width=1cm, minimum height=0.5cm]
\tikzstyle{legend_fw}=[rectangle, rounded corners, very thin,
                           black, fill= asuorange!8, draw, text=black]

\usepackage{etoolbox}
\makeatletter
\patchcmd{\@makecaption}
  {\scshape}
  {}
  {}
  {}
\makeatletter
\patchcmd{\@makecaption}
  {\\}
  {.\ }
  {}
  {}
\makeatother
\def\tablename{\bf Table}

\begin{document}

\bibliographystyle{IEEEtran}

\newcommand{\papertitle}
{
Curvature-aided Incremental Aggregated Gradient Method
}

\newcommand{\paperabstract}{
...
}


\ifplainver


    \title{\papertitle}

    \author{
    Hoi-To Wai, Wei Shi, Angelia Nedi\'c and Anna Scaglione
    \thanks{This work  is supported by NSF CCF-BSF 1714672. The authors are with the School of Electrical, Computer and Energy Engineering, Arizona State University, Tempe, AZ 85281, USA. E-mails: \texttt{\{htwai,wshi36,Angelia.Nedich,Anna.Scaglione\}@asu.edu}.}
    }

    \maketitle

\else
    \title{\papertitle}

    \ifconfver \else {\linespread{1.1} \rm \fi

    \author{\IEEEauthorblockN{Hoi-To Wai\IEEEauthorrefmark{1}, Wei Shi\IEEEauthorrefmark{1}, Angelia Nedi\'c\IEEEauthorrefmark{1} and Anna Scaglione\IEEEauthorrefmark{1}}
\IEEEauthorblockA{\IEEEauthorrefmark{1}
School of Electrical, Computer and Energy Engineering, Arizona State University, Tempe, AZ 85281, USA.\\
Emails: \texttt{\{htwai,wshi36,angelia.nedich,Anna.Scaglione\}@asu.edu}}
         }

    \maketitle

    \ifconfver \else
        \begin{center} \vspace*{-2\baselineskip}
        \end{center}
    \fi

    \ifconfver \else \IEEEpeerreviewmaketitle} \fi

 \fi
 
     \begin{abstract}
We propose a new algorithm for finite sum optimization
     which we call the curvature-aided incremental aggregated
     gradient ({\sf CIAG}) method. 
     Motivated by the problem of training a classifier for a $d$-dimensional problem, 
     where the number of training data is $m$ and $m \gg d \gg 1$, 
     the {\sf CIAG} method seeks to accelerate
     incremental aggregated gradient ({\sf IAG}) methods using aids from the
     curvature (or Hessian) information, while avoiding the
     evaluation of matrix inverses required by the
     incremental Newton ({\sf IN}) method. 
     Specifically, our idea is to exploit
     the incrementally aggregated Hessian matrix to trace the
     full gradient vector at every incremental step, therefore
     achieving an improved linear convergence rate over the
     state-of-the-art {\sf IAG} methods. 
     For strongly convex problems, 
     the fast linear convergence rate requires the objective function
     to be close to quadratic, or the initial point to be close to optimal solution. 
     Importantly, we show that running \emph{one} iteration of
     the {\sf CIAG} method yields the same improvement to the optimality gap
     as running one iteration of the \emph{full gradient} method,
     while the complexity is ${\cal O}(d^2)$ for {\sf CIAG} and ${\cal O}(md)$ 
     for the full gradient.
     Overall, the {\sf CIAG} method strikes a balance
     between the high computation complexity
     incremental Newton-type methods and the slow {\sf IAG}
     method. 
     Our numerical results support the theoretical findings and 
     show that the {\sf CIAG} method
     often converges with much fewer iterations than {\sf IAG},
     and requires much shorter running time than {\sf IN} when the problem
     dimension is high.\end{abstract}

\ifplainver
\else
    \begin{IEEEkeywords}\vspace{-0.0cm}
       incremental gradient method, linear convergence, Newton method, empirical risk minimization
    \end{IEEEkeywords}
\fi

\ifconfver \else
    \ifplainver \else
        \newpage
\fi \fi

\section{Introduction}
In this paper we consider the unconstrained optimization problem
whose objective is a finite sum of functions, \ie
\beq \label{eq:opt}
\min_{ \prm \in \RR^d }~F(\prm) \eqdef \sum_{i=1}^m f_i ( \prm ) \eqs,
\eeq
where each $f_i : \RR^d \rightarrow \RR$ 
is a convex and twice continuously differentiable function
and $\prm \in \RR^d$ is the parameter variable.
We shall refer to each $f_i (\prm)$ as a component function and 
$F(\prm)$ as the sum objective function.
The problem above is motivated by many practical machine learning applications. 
For instance, consider training a support vector machine 
(SVM) system with the data set $\{ ({\bm x}_i, y_i) \}_{i=1}^m$.
To cast the training problem in the form of \eqref{eq:opt},  
each of the function $f_i (\prm)$ can be set
as a loss function
$\ell ( \prm; {\bm x}_i, y_i )$ 
which measures the statistical mismatch between the parameter $\prm$ 
and the data tuple $( {\bm x}_i, y_i )$. 

We focus on a large-scale optimization setting where $m \gg d \gg 1$, 
considering the prototypical situation of the so called ``big data'' challenge
in machine learning. 
In fact, in the SVM classifier case this situation arises when we use 
a large amount of training data ($m \gg 1$) 
on a problem with a large number of features ($d \gg 1$).  
To cope with these challenges, a popular approach is to apply
\emph{incremental methods}~\cite{bertsekas2011incremental} to
Problem \eqref{eq:opt}. Compared
to conventional methods, such as gradient or Newton methods, which require
accessing to \emph{all} $m$  
component functions at every iteration, incremental 
methods access to only \emph{one} of the component functions at 
each iteration and, therefore,
the complexity per iteration is independent of $m$. 
The prior art on incremental methods 
have focused on the \emph{gradient}-type methods
which approximate the gradient method. Related work include the 
celebrated stochastic gradient descent method~\cite{robbins1951stochastic} 
and its accelerated variants 
\cite{roux2012stochastic,johnson2013accelerating,defazio2014saga}. 
The deterministic counterpart of these methods have been studied, e.g., 
in~\cite{blatt2007convergent,vanli2016stronger,gurbuzbalaban2017convergence,mokhtari2016surpassing},
while incremental subgradient methods for nonsmooth problems have been investigated
in~\cite{nedic2000,nedic2001a,nedic2001b}.
Even though the gradient algorithms are shown to converge linearly for strongly convex
problems, the convergence rate
is typically slow. In fact, it was shown in \cite{vanli2016stronger} that 
the linear convergence rate of the incremental
aggregated gradient ({\sf IAG}) method is $1 - {\cal O}(1/Qm)$ 
where $Q$ is the condition 
number of the objective function $F$ in~\eqref{eq:opt}, 
and an improved {\sf IAG} method was studied in \cite{mokhtari2016surpassing}. 
To accelerate the convergence, there has been recently renewed interest in 
studying 
incremental \emph{Newton}(\emph{-type}) methods. 
For example, the incremental Newton ({\sf IN}) 
method in \cite{rodomanov2016superlinearly}
and the incremental quasi Newton method ({\sf IQN}) 
in \cite{mokhtari2017iqn}. While these algorithms are shown to
achieve superlinear local convergence for strongly convex problems, 
the computational complexity is usually high,
the algorithm involves computing the Hessian inverses.

This paper proposes a new algorithm called curvature-aided incremental 
aggregated gradient ({\sf CIAG}) method which is a first order method based
on gradient descent. To accelerate convergence, the method exploits
\emph{curvature information} (Hessian) to aid in tracing the full gradient at \emph{every}
iteration using only incremental information. 
We first show that the {\sf CIAG} is globally convergent with 
a sufficiently small step size when the objective function is strongly convex. 
Furthermore, when the objective function $F(\prm)$ is quadratic-like or when the {\sf CIAG}
method is initialized at a point close to the optimal solution, 
we show that each \emph{incremental step} of the {\sf CIAG} 
method can asymptotically achieve a similar linear rate as applying an iteration of
the \emph{full gradient} step. In other words, 
the method converges to an optimal solution
of \eqref{eq:opt} at an equivalent rate of running $m$ gradient steps 
after one cycle of accessing the component functions. 
We also suggest an adaptive
step size rule that can conceptually attain the accelerated convergence rate. 
This results in an efficient algorithm with fast
convergence at a low complexity.
We show a comparison of {\sf CIAG} to the state-of-the-art methods in Table~\ref{tab:com}. 
 
\begin{table}[t]
\begin{center}
\renewcommand\baselinestretch{1.2}\selectfont
\begin{tabular}{c||c|c|c} 
\hline
& Storage & Computation & ${\ds \lim_{k \rightarrow \infty}} \frac{\| \prm^{k+1} - \prm^\star \|^2}{\| \prm^{k} - \prm^\star \|^2}$  \\
\hline
\hline 
{\sf FG} & ${\cal O}(d)$ & ${\cal O}( md )$ & $1 - 4Q/(Q+1)^2$ \\
\hline
{\sf IG} \cite{bertsekas2011incremental} & ${\cal O}(d)$ & ${\cal O}(d)$ & 1,~\ie sub-linear \\
\hline
{\sf IAG} \cite{vanli2016stronger} & ${\cal O}(md)$ & ${\cal O}(d)$ & $1-{\cal O}(1/Qm)$~$^\dagger$ \\
\hline
{\sf IQN} \cite{mokhtari2017iqn} & ${\cal O}(md^2)$ & ${\cal O}(d^2)$ & 0,~\ie super-linear \\ 
\hline
{\sf IN} \cite{rodomanov2016superlinearly} & ${\cal O}(md)$ & ${\cal O}(d^3)$ & 0,~\ie super-linear \\
\hline
\hline
{\sf CIAG} (Proposed) & ${\cal O}(md)$ & ${\cal O}(d^2)$ & $1- 4Q/(Q+1)^2$~$^\ddagger$ \\
\hline
\end{tabular}\vspace{.2cm}
\end{center}
\caption{Comparison of the storage \& computation
complexities, convergence speed of different methods. 
The computation complexity is stated in the per \emph{iteration} sense, which
refers to the computational cost of
the full gradient for {\sf FG}, and the incremental 
gradient/Hessian for the incremental methods. 
The last column is the 
local linear convergence rate 
and $Q = L / \mu$ is the condition number of \eqref{eq:opt}. \\[.2cm]
{\footnotesize $^\dagger$Note that \cite{vanli2016stronger}
analyzed the convergence rate of {\sf IAG} 
in terms of the optimality gap of objective value, showing that
$F(\prm^k) - F(\prm^\star) \leq (1 - 1 / (49Qm))^k (F(\prm^0) - F(\prm^\star))$,
where $\prm^\star$ is an optimal solution.\\[.2cm]
$^\ddagger$The $(1- 4Q/(Q+1)^2)$ rate of {\sf CIAG} is achieved using an adaptive step size  described in Section~\ref{sec:ciag_d}.}} \label{tab:com} \vspace{-0.5cm}
\end{table}
 
\subsection{Notations and Preliminaries}
For any $d \in \NN$, we use the notation $[d]$ to refer to the set $\{1,...,d\}$.
We use boldfaced lower-case letters to denote vectors and boldfaced upper-case letters to denote matrices.
The positive operator $(x)_+$ denotes $\max\{0,x\}$.
For a vector ${\bm x}$ (or a matrix ${\bm X}$),
the notation $[ {\bm x} ]_i$ (or $[{\bm X}]_{i,j}$) denotes its $i$th element (or $(i,j)$th element).
For some positive finite constants $C_1,C_2,C_3,C_4$ where $C_3 \leq C_4$,
and non-negative functions $f(t), g(t)$,
the notations $f(t) = {\cal O}( g(t) )$, $f(t) = \Omega( g(t) )$, $f(t) = \Theta( g(t) )$
indicate $f(t) \leq C_1 g(t)$, $f(t) \geq C_2 g(t)$, $C_3 g(t) \leq f(t) \leq C_4 g(t)$, 
respectively.
Unless otherwise specified, $\| \cdot \|$ denotes the standard Euclidean norm.

\section{The CIAG Method}
We develop the {\sf CIAG} method starting from the classical 
gradient method. We shall call the latter 
as \emph{full gradient method} ({\sf FG}) from now on to distinguish 
it from incremental gradient methods. The {\sf FG} method applied to \eqref{eq:opt}
can be described by the recursion: at iteration $k \in \NN$,
\beq
\prm^{k+1} = \prm^k - \gamma \grd F(\prm^k) = \prm^k - \gamma \sum_{i=1}^m \grd f_i (\prm^k) \eqs,
\eeq
where $\gamma > 0$ is a step size. 
Notice that evaluating the update above requires accessing the 
gradients of \emph{all} component functions at every iteration. This 
is expensive in terms of the computation cost as $m \gg 1$. 


The {\sf CIAG} method adopts the \emph{incremental update} paradigm
to address the  complexity issue above, \ie at the $k$th iteration, we
restrict the algorithm's access to only \emph{one} component
function, say the $i_k$th function $f_{i_k}(\prm)$.
As desired, the per-iteration computation 
cost will become independent of $m$.
However, 
this also implies that the \emph{exact gradient} vector $\grd F(\prm^k)$ is no longer 
available
since the rest of the component functions are not accessible. 
Our idea is to apply the following Taylor's approximation:
\beq \label{eq:taylor}
\grd f_j ( \prm^k ) \approx \grd f_j ( \prm' ) + \grd^2 f_j (\prm' ) ( \prm^k - \prm' ) \eqs.
\eeq
It suggests that even when the $j$th function is not available at the current iteration $k$,
its gradient $\grd f_j( \prm^k)$ can still be approximated 
using the historical gradients/Hessians.
To this end, let us model the {\sf CIAG} method using a delayed gradient/Hessian setting
similar to \cite{gurbuzbalaban2017convergence}. 
Let us define $\tau_j^k \in \NN$ as the \emph{iteration number}
in which the {\sf CIAG} method has last accessed to $f_j (\cdot)$ prior
to iteration $k$, \ie
\beq
\tau_j^k \eqdef \max \{ k' ~:~i_{k'} = j,~k' \leq k \} \eqs.
\eeq
Note that $\tau_{i_k}^k = k$. 
We assume $m \leq K < \infty$ such that
\beq \label{eq:finite}
\max\{0, k - K \} \leq \tau_j^k \leq k,~j=1,...,m \eqs,
\eeq
and at iteration $k$, the vectors of the following historical iterates 
are available:
\beq \label{eq:store}
\{ \prm^{\tau_j^k} \}_{j=1}^m \eqs.
\eeq
The exact gradient can then be approximated by the following 
Taylor's approximation: 
\beq \label{eq:grd_app}
\tilde{\bm g}^k \eqdef \sum_{i=1}^m \Big( 
\grd f_i ( \prm^{\tau_i^k} ) + \grd^2 f_i( \prm^{\tau_i^k} ) ( \prm^k - \prm^{\tau_i^k} ) \Big) \eqs,
\eeq
where $\tilde{\bm g}^k \approx \sum_{i=1}^m \grd f_i (\prm^k)$. 
We remark that the \emph{incremental
aggregated gradient} ({\sf IAG}) method takes almost the same 
form of update as {\sf CIAG}, \ie the {\sf IAG} method uses
\beq
\sum_{i=1}^m \grd f_i (\prm^k) \approx \sum_{i=1}^m \grd f_i ( \prm^{\tau_i^k} ) \eqs.
\eeq 

\algsetup{indent=0.8em}
\begin{algorithm}[t]
\caption{{\sf CIAG} Method.}\label{alg:ciag}
  \begin{algorithmic}[1]
  \STATE \textbf{Input}: Initial point $\prm^1 \in \RR^d$.
  \STATE Initialize the vectors/matrices:
  \beq
  \prm_i \leftarrow \prm^1,~i=1,...,m,~ {\bm b}^0 \leftarrow {\bm 0},~{\bm H}^0 \leftarrow m {\bm I} \eqs. \vspace{-.4cm}
  \eeq
  \FOR {$k=1,2,\dots$}
  \STATE \label{ciag:sel} Select $i_k \in \{1,...,m\}$, e.g., $i_k = (k~{\rm mod}~m) + 1$.
   \STATE \label{ciag:sums} Update the vector and matrix ${\bm b}^k, {\bm H}^k$ as:
   \beq 
   \begin{split}
   {\bm b}^{k} & = {\bm b}^{k-1} - \grd f_{i_k}( \prm_{i_k} ) + \grd f_{i_k} (\prm^k) \\
   & \hspace{0.8cm} + \grd^2 f_{i_k}( \prm_{i_k} ) \prm_{i_k} - \grd^2 f_{i_k} (\prm^k) \prm^k \eqs, \\
   {\bm H}^{k} & = {\bm H}^{k-1} - \grd^2 f_{i_k}( \prm_{i_k} ) + \grd^2 f_{i_k} (\prm^k) \eqs.\\[-.1cm]
   \end{split}
   \eeq
   \STATE \label{ciag:upd} Compute the {\sf CIAG} update:
   \beq \label{eq:ciag_imp} \textstyle
   \prm^{k+1} = \prm^k - \gamma \big( {\bm b}^k + {\bm H}^k \prm^k   \big) \eqs. \vspace{-.4cm}
   \eeq
   \STATE \label{ciag:mem} Update the parameter stored in memory $\prm_{i_k} \leftarrow \prm^k$. 
\ENDFOR
\STATE \textbf{Return}: an approximate solution to \eqref{eq:opt}, $\prm^{k+1}$.
  \end{algorithmic}
\end{algorithm}

From \eqref{eq:grd_app}, we note that one needs to aggregate \emph{(i)} 
the incremental gradient $\grd f_i ( \prm^{\tau_i^k} )$, \emph{(ii)} the incremental 
Hessian-iterate product $\grd^2 f_i ( \prm^{\tau_i^k} ) \prm^{\tau_i^k}$ and \emph{(iii)} the 
incremental Hessian $\grd^2 f_i( \prm^{\tau_i^k} )$, 
in order to compute an {\sf CIAG} update. These can be achieved 
efficiently using the procedure outlined in Algorithm~\ref{alg:ciag}\footnote{In the pseudo 
code, we use $\prm_i,~i=1,...,m$
to denote the variable that we use to store $\prm^{\tau_i^k}$, \ie the historical 
parameter at the iteration when the $i$th component function is last accessed.}.
In particular, the updates in line~\ref{ciag:sums} essentially swap out the previously 
aggregated gradient/Hessian information and replace them with the ones computed
at the current iterate for the $i_k$th component function; and line~\ref{ciag:mem}
keeps track of the historical iterates to ensure that $\prm^{\tau_{i_k}^k}$ is stored.
It can be verified that the term, ${\bm b}^k + {\bm H}^k \prm^k$, 
inside the bracket of \eqref{eq:ciag_imp}
is equivalent to the right hand side of \eqref{eq:grd_app}. 
Both forms of the {\sf CIAG} update are presented as the form in \eqref{eq:ciag_imp} 
gives a more efficient implementation, while \eqref{eq:grd_app} is more tamable for 
analysis. 
Furthermore,  for the special case with a cyclic 
selection rule for the component functions, the sequence
of $\tau_i^k$ satisfies \eqref{eq:finite} with $K = m$. 

Lastly, we comment on the complexity of the {\sf CIAG} method. 
First let us focus on the computation complexity --- 
from Line~\ref{ciag:sums} to \ref{ciag:upd} in Algorithm~\ref{alg:ciag}, we 
observe that the {\sf CIAG} method requires a computation 
complexity\footnote{We have neglected the computation complexity of evaluating
$\grd f_i (\prm^k)$ and $\grd^2 f_i(\prm^k)$ as they are often problem dependent.
The difference that it makes is minimal as long as the algorithms are \emph{incremental},
\ie does not require accessing a large number of these gradient/Hessian.} 
of ${\cal O}(d^2)$ (due to the matrix-vector multiplication) per iteration;
meanwhile, the {\sf IAG} method requires ${\cal O}(d)$, the 
{\sf IQN} method requires ${\cal O}(d^2)$ and the {\sf IN} method
requires ${\cal O}(d^3)$ per iteration.
Secondly, the storage requirement for {\sf CIAG}, {\sf IAG} and {\sf IN} methods 
are the same, \ie ${\cal O}(md)$ is needed in storing \eqref{eq:store}, 
while the {\sf IQN} method requires ${\cal O}(md^2)$ to store the historical quasi-Hessians.
The above analysis shows that the {\sf CIAG} method provides a tradeoff 
between the slow convergence in 
{\sf IAG} method and the high complexity of {\sf IN} or {\sf IQN} method. 
These comparisons are summarized in Table~\ref{tab:com}. 

\section{Convergence Analysis}
In this section we analyze the convergence of the {\sf CIAG} method.
We focus on strongly convex problems, 
where the {\sf CIAG} method is shown to converge linearly.
To proceed with our analysis, 
we state a few required assumptions on the component functions $f_i (\prm )$ and 
sum function $F (\prm )$.
\begin{Assumption}\label{ass:cts_h}
The Hessian of each of 
the component function $f_i (\prm)$ is $L_{H,i}$-Lipschitz. In other words, for all $\prm', \prm \in \RR^d$, 
\beq \begin{split}
& \| \grd^2 f_i ( \prm ) - \grd^2 f_i (\prm' ) \| \leq L_{H,i} \| \prm - \prm' \| \eqs.
\end{split}
\eeq
\end{Assumption}
Note that if $f_i(\prm)$ is a quadratic function, then $L_{H,i} = 0$. 
Furthermore, we define $L_H \eqdef \sum_{i=1}^m L_{H,i}$
as the Hessian Lipschitz constant for $F(\prm)$.
\begin{Assumption}\label{ass:cts}
The gradient of the sum function $F(\prm)$ is $L$-Lipschitz. In other words,
for all $\prm', \prm \in \RR^d$, 
\beq \begin{split}
& \| \grd F ( \prm ) - \grd F (\prm' ) \| \leq L \| \prm - \prm' \| \eqs. \\
\end{split}
\eeq
\end{Assumption}
We remark that, if the $i$th component function has a $L_i$-Lipschitz gradient, 
then the Lipschitz constant for the gradient of $F(\prm)$ can be 
bounded as $L \leq \sum_{i=1}^m L_i$.
\begin{Assumption}\label{ass:strcvx}
The sum function $F (\prm)$ is $\mu$-strongly convex, $\mu > 0$, \ie for all $\prm', \prm \in \RR^d$, 
\beq
F( \prm' ) \geq F( \prm ) + \langle \grd F(\prm), \prm' - \prm \rangle + \frac{\mu}{2} \| \prm' - \prm \|^2 \eqs.
\eeq
\end{Assumption}
Under Assumption~\ref{ass:strcvx}, a unique optimal solution 
to problem~\eqref{eq:opt} exists and it is denoted by $\prm^\star$. 

As a matter of fact, objective functions satisfying the above assumptions are common in  
machine learning. For example, consider the logistic loss function
corresponding to a data tuple $(y_i,{\bm x}_i)$, where $y_i \in \{ \pm 1 \}$
is the data label and ${\bm x}_i \in \RR^d$ is the associated feature vector, the
loss function is given as:
\beq \label{eq:log}
\ell ( \prm ; ( y_i, {\bm x}_i ) ) = \frac{1}{\rho} \log ( 1 + e^{-y_i \langle {\bm x}_i, \prm \rangle} ) + \frac{1}{2} \| \prm \|^2 \eqs,
\eeq
where the latter term is a standard $\ell_2$ regularizer
and $\rho > 0$ controls the strength of regularization.
Taking $f_i ( \prm ) = \ell( \prm; (y_i, {\bm x}_i) )$ in problem \eqref{eq:opt}, 
it can be verified that problem~\eqref{eq:opt} satisfies 
Assumption~\ref{ass:cts_h} to \ref{ass:strcvx} with the following
Lipschitz constants: 
\beq \label{eq:log_const}
L = \frac{1}{\rho} \| {\textstyle \sum_{i=1}^m {\bm x}_i {\bm x}_i^\top} \|_2 + m,~L_{H,i} = \frac{1}{\rho} \| {\bm x}_i {\bm x}_i^\top \|_2 \eqs,
\eeq
and the strongly convex constant of $\mu = m$. 
We observe that if $\rho$ is large, then the Hessian's Lipschitz 
constant $L_{H,i}$ is small. 


We next show that the {\sf CIAG} method converges linearly for strongly convex 
problems. 
The following holds:
\begin{Theorem} \label{thm:scvx}
Under Assumption~\ref{ass:cts_h}, \ref{ass:cts} and \ref{ass:strcvx}. Let 
$\dst{k} \eqdef \| \prm^k - \prm^\star \|^2$ where $\prm^k$ is the $k$th
iterate generated by {\sf CIAG} and $\prm^\star$ is the optimal solution to \eqref{eq:opt}. 
Fix $s \in \NN$ as an arbitrary integer and $\epsilon > 0$, 
if the step size $\gamma$ satisfies:
\beq \label{eq:step}
\begin{split} & \gamma \leq \epsilon + \min\Big\{ \frac{2}{\mu + L}, \\
& \frac{1}{2K} \sqrt{ \frac{\mu L}{L_H (L^2 \dst{s}^{1/2} +16L_H^2 \dst{s}^{3/2} ) (\mu+L) } },\\
& \Big( \frac{1}{8K^4} \frac{ \mu L }{ L_H^2 (L^4 \dst{s} + 256L_H^4 \dst{s}^3 ) (\mu+L) } \Big)^{1/5}  \Big\} \eqs.
\end{split}
\eeq 
then the sequence $\{ \dst{k} \}_{k \geq s}$ 
converges linearly as:
\beq \label{eq:linear1}
\dst{k} \leq (1-\epsilon)^{\lceil (k-s)/(2K+1) \rceil} \dst{s} ,~\forall~ k \geq s \eqs.
\eeq
Moreover, asymptotically the linear rate can be improved to:
\beq \label{eq:rate}
\begin{split}
& \lim_{k \rightarrow \infty} \frac{ \dst{k+1} }{ \dst{k} } \leq 1 - 2 \gamma \frac{ \mu L }{ L + \mu } \eqs,
\end{split}
\eeq
\ie as $k \rightarrow \infty$, 
the {\sf CIAG} method converges linearly at the same rate of {\sf FG} using
the step size $\gamma$.
\end{Theorem}
The next subsection provides a proof sketch of Theorem~\ref{thm:scvx}\ifplainver
.
\else
\footnote{The
complete proof can be found in an online appendix, available at: \texttt{\url{http://www.public.asu.edu/~hwai2/pdf/CIAG_Proof.pdf}}.}.
\fi

We observe that the convergence rate of {\sf CIAG} in \eqref{eq:rate}
is equivalent to that of {\sf FG} with the same step size $\gamma$. 
Importantly, to achieve the same worst-case decrement of the distance to the optimal solution,
the {\sf CIAG} method only requires access to \emph{one} component function
at each iteration, while the {\sf FG} method requires access to all the $m$ component
functions. 
Furthermore, if the objective function is `quadratic'-like, \ie when $L_H$ is small,
then one can take $\gamma \approx 2 / (\mu+L)$, which is the same
maximum allowable step size for the
{\sf FG} method \cite{bertsekas1999nonlinear}. 
In the above case, the {\sf CIAG} method
has a linear convergence rate of
\beq
1 - \frac{4 \mu L}{(\mu+L)^2} = 1 - \frac{4Q}{(Q+1)^2}  \eqs,
\eeq
where $Q = L / \mu$ is the condition number of \eqref{eq:opt}.

Theorem~\ref{thm:scvx} 
shows that the {\sf CIAG} method is \emph{globally convergent} when an appropriate, fixed
step size is chosen. This is in contrast to other curvature information
based methods such as {\sf IQN}, which only have 
local convergence guarantee. One of the main reasons 
is that the {\sf CIAG} method is developed as an approximation to
the classical {\sf FG} method, whose global convergence is established 
with less restrictions.

\subsection{Proof Sketch of Theorem~\ref{thm:scvx}}
As opposed to the {\sf IAG} method, the {\sf CIAG} method applies 
a first-order Taylor's approximation to the gradient vector. By utilizing 
Assumption~\ref{ass:cts_h} (Lipschitz smoothness of the Hessian), 
it can be shown that the gradient error is bounded by:
\beq
\| \grd F(\prm^k) - \tilde{\bm g}^k \| = {\cal O}( \gamma^2  m^2 \max_{ k' \in [(k-K)_+,k] }  \| \prm^{k'} - \prm^\star \|^2 )\eqs.
\eeq
We observe that the \emph{norm} of the error is bounded by a \emph{squared norm}
of the optimality gap $\| \prm^{k'} - \prm^\star \|^2$. The latter decays quickly 
when the optimality gap is close to zero. 
Consequently, studying the iteration of the optimality gap leads us to
the following inequality:
\beq \label{eq:ciag_sketch} \begin{split}
\| \prm^{k+1} - \prm^\star \|^2 & \leq \Big( 1 - 2 \gamma \frac{\mu L}{L + \mu} \Big) \|
\prm^{k} - \prm^\star \|^2 \\
& \hspace{-1cm} + {\cal O} \Big( \gamma^3 m^2 \max_{ k' \in [(k-2K)_+,k] } \| \prm^{k'} - \prm^\star \|^3 \Big) \eqs,
\end{split}
\eeq
notice that difference in the power of the term $\| \prm^k - \prm^\star \|$ on the right hand side
and the fact that the expression $(1- 2\gamma \mu L / (\mu+L))$ is the linear rate
obtained for the {\sf FG} method \cite{nesterov2013introductory}. 
In other words, the optimality gap at the $(k+1)$th iteration 
is bounded by the sum of a term that decays with a linear rate 
of $(1- 2\gamma \mu L / (\mu+L))$ and delayed error terms 
of higher order. 

The observation above prompts us to study the following general 
inequality system with the non-negative sequence:
consider $\{ R(k) \}_{k\geq 0}$ 
satisfying the inequality:
\beq \label{eq:ineq2} \textstyle
R(k+1) \leq p R(k) + \sum_{j=1}^J q_j \max_{ k' \in {\cal S}_j^k } R(k')^{\eta_j}  \eqs,
\eeq
where $0\leq p<1$, $q_j \geq 0$, $\eta_j > 1$ and ${\cal S}_j^k \subseteq [(k-M+1)_+,k]$ for all $j$
with some $J, M < \infty$. We have
\begin{Lemma} \label{lem:2}
For some $p \leq \delta < 1$, if 
\beq \label{eq:cond2} \textstyle
p + \sum_{j=1}^J q_j R(0)^{\eta_j - 1} \leq \delta < 1 \eqs,
\eeq 
then (a) $\{R(k)\}_k$ converges linearly for all $k$ as
\beq \label{eq:lem1}
R(k) \leq \delta^{ \lceil k / M \rceil } \cdot R(0),~\forall~k \geq 0 \eqs, 
\eeq 
and (b) the rate is accelerated to $p$ asymptotically, 
\beq \label{eq:lem2}
\begin{split}
& \lim_{k \rightarrow \infty} R(k+1) / R(k) \leq p \eqs.
\end{split}
\eeq
\end{Lemma}
Observe that \eqref{eq:ciag_sketch} fits into the requirement 
of \eqref{eq:ineq2} with $R(k) = \| \prm^k - \prm^\star \|^2$, $M=2K$ 
and $p = 1 - 2\gamma \mu L / (L + \mu)$. 
The claims in Theorem~\ref{thm:scvx} thus follow by 
applying Lemma~\ref{lem:2} and 
carefully characterizing the constants. 

\subsection{Linear Convergence Rate of {\sf CIAG} Method} \label{sec:ciag_d}
We note that the linear convergence rate of {\sf CIAG} hinges on the 
choice of step size $\gamma$ as specified in \eqref{eq:step},
where the latter depends on the number of parameters in the optimization problem.
This section discusses the relationship between the 
convergence rate and the choice of step size $\gamma$ under different settings. \vspace{.2cm}

\noindent \textbf{Worst-case convergence rate with a constant step size ---} 
We now compare the (worst-case) convergence rate of the {\sf CIAG} 
method with other incremental methods when using a constant step size.
To proceed, let us define the following constants:
\beq
Q \eqdef \frac{L}{\mu},~Q_H \eqdef \frac{L}{L_H} = \frac{\mu}{L_H} Q \eqs,
\eeq
which correspond to the condition number of \eqref{eq:opt} and
the ratio between the gradient's and Hessian's Lipschitz constant.
Substituting the constants above into the expression in \eqref{eq:step} of the
{\sf CIAG}'s step size and that the second term 
inside $\min\{\cdot\}$ is dominant when $K$ is large, 
the step size has to satisfy:
\beq
\gamma = \frac{1}{2K} \sqrt{ \frac{Q_H}{ (L^2 \dst{0}^{1/2} +16L_H^2 \dst{0}^{3/2} ) (Q+1) } }  - \epsilon \eqs,
\eeq
The linear convergence rate achieved at $k \rightarrow \infty$ is 
$1-p$, where $p$ is approximately:
\beq
\begin{split}
p & \approx \frac{L}{K ( Q+1 )}\sqrt{ \frac{Q_H}{ (L^2 \dst{0}^{\frac{1}{2}} +16L_H^2 \dst{0}^{\frac{3}{2}} ) (Q+1) } } \\
& = \frac{1}{K (Q+1)} \sqrt{ \frac{Q_H}{Q+1} } \sqrt{ \frac{1}{V(0)^{\frac{1}{2}} + 16 V(0)^{\frac{3}{2}} / Q_H^2} } \eqs.
\end{split}
\eeq
It is instructive to compare the rate rate to the best known rate of the {\sf IAG} method
from \cite{vanli2016stronger}. In particular, \cite[Theorem 3.4]{vanli2016stronger} shows that the linear convergence rate 
of {\sf IAG} is given by $1-p'$, with
\beq
p' = \frac{1}{K(Q+1)} \frac{1}{49} \eqs.
\eeq
The comparison above shows that the {\sf CIAG} method has at least the same scaling
as the {\sf IAG} method, as well as the {\sf SAG} method, \ie the stochastic 
counterpart of the {\sf IAG} method. 

Moreover, we observe that when 
\beq \label{eq:compare}
\sqrt{ \frac{Q_H}{Q+1} } \sqrt{ \frac{1}{V(0)^{\frac{1}{2}} + 16 V(0)^{\frac{3}{2}} / Q_H^2} } > \frac{1}{49} \eqs,
\eeq
then the {\sf CIAG} method enjoys a faster convergence rate than the {\sf IAG} method. 
Notice that this depends on the ratio $Q_H / Q = \mu / L_H$ and the 
initial distance to optimal solution $\dst{0}$ such that \eqref{eq:compare}
holds when $L_H$ is small or when $\dst{0}$ is small. 
We remark that the 
analysis above is a representative of a worst case scenario for the {\sf CIAG} method.
In practice, we find that {\sf CIAG} method is convergent for most cases when the largest 
possible step size $2 / (\mu+L)$ is chosen.
Furthermore, in the following, we argue that the {\sf CIAG} algorithm 
can be accelerated significantly using an \emph{adaptive step size} rule. 
\vspace{.2cm}

\noindent \textbf{Achieving {\sf FG}'s rate with one incremental step ---}
An important insight from Theorem~\ref{thm:scvx} is that the maximum allowable 
step size $\gamma$ 
is inversely proportional to the initial distance to optimal solution, $\dst{s}$. 
In particular, it is possible to set the step size to the maximum allowable
value $\gamma = 2 / (\mu+L)$
when the initial point is sufficiently close to the optimum.
In fact, for large $K$, whenever the optimality gap $\dst{s}$ satisfies
\beq \label{eq:sat}
\dst{s}^{\frac{1}{2}} + 16 \frac{\dst{s}^{\frac{3}{2}}}{Q_H^2} < \frac{1}{16K^2} \frac{Q_H}{Q+1} \Big( \frac{Q+1}{Q} \Big)^2 \eqs,
\eeq
then the {\sf CIAG} method is allowed to take the maximum allowable step size
$2 / (\mu+L)$ at the iteration numbers $k \geq s$.

For the given condition numbers $Q,Q_H$ and an initial step size 
choice that satisfies \eqref{eq:step} at $s=0$, 
Theorem~\ref{thm:scvx} shows that 
Eq.~\eqref{eq:sat} can be satisfied when $s$ is large,
since the optimality gap $\dst{s}$ also decreases 
to zero linearly with the iteration number $s$.
This suggests that the step size $2 / (\mu+L)$
is allowable once we run the {\sf CIAG} method for long enough time. 
Note that this holds regardless of the `quadratic'-ness, $L_H$, of the objective function. 
As an heuristic, we may apply a time-varying step size 
that increases to $2 / (\mu+L)$.
Designing a globally converging step size scheme 
and analyzing its property
are left as future work topics.

\section{Numerical Experiments}
In this section, we compare the performance of several
incremental methods. 
For a fair comparison, in the following
we shall concentrate on testing 
the incremental methods with the deterministic, cyclic component selection rule
similar to Line~\ref{ciag:sel} of Algorithm~\ref{alg:ciag}, \ie 
we shall not test the stochastic variants of the incremental methods, e.g.,
{\sf SAG} \cite{roux2012stochastic} or {\sf SAGA} \cite{defazio2014saga}. 
We focus on the problem of training an SVM using the logistic loss function 
given in \eqref{eq:log}. 
For the logistic loss function \eqref{eq:log}, we set 
$\rho = 1/m$. 
Note that as analyzed in \eqref{eq:log_const}, the 
condition number $Q$ increases with $m$ in general. 

Our first set of experiments considers training an SVM with synthetic data ---
we first generate $\prm_{\sf true} \in \RR^d$ as a random vector with 
each element distributed 
independently as ${\cal U}[-1,1]$; 
then each of the feature vector ${\bm x}_i \in \RR^d$ 
is generated as 
${\bm x}_i = [ \tilde{\bm x}_i;~1 ]$ where $\tilde{\bm x}_i \sim {\cal U}[-1,1]^{d-1}$ 
and the label $y_i$ is 
computed as $y_i = {\rm sign}( \langle {\bm x}_i, \prm_{\sf true} \rangle )$. 
We remark that the set up described is equivalent to an SVM with bias 
term, which is given by the last element of $\prm_{\sf true}$.

To set up the benchmark algorithms, 
the step sizes for both {\sf IN} and {\sf IQN} are both set as $\gamma = 1$
to attain the fastest convergence (at superlinear rate) 
even though it may not guarantee global convergence. 
For the {\sf IAG} method, we have \emph{optimized} 
the step size and we set $\gamma = 50 / (mL)$, where $L$
is the Lipschitz constant for $\grd F(\prm)$
that is computed as 
$m + m \| \sum_{i=1}^m {\bm x}_i {\bm x}_i^\top \|_2$. 
We also test the incremental gradient ({\sf IG}) method 
using a vanishing step size $\gamma_k = 1 / ( \lceil k / m \rceil L )$.
For the {\sf CIAG} method, we set $\gamma = 1 / L$.

\begin{figure}[t]
\centering
\ifplainver
\resizebox{.7\linewidth}{!}{\resizebox{.98\linewidth}{!}
{ \sf 
\begin{tikzpicture}
\pgfplotsset{ scale only axis,
    width=0.55\textwidth,height=0.3\textwidth,
    grid=both,grid style={line width=.01pt, draw=gray!30},
    legend cell align=left, legend style={legend pos=south east,font=\large, yshift = 0.4cm},
    xlabel={\large Effective Pass ($k/m$)},
    enlarge x limits=0.05,
    enlarge y limits=0.05,
    xmin = 1.5, xmax = 25,
}

\pgfplotsset{every tick label/.append style={font=\large}}

\begin{semilogyaxis}[ylabel shift = -0.1cm, ylabel={\large Opt.~Gap $F(\prm^k) - F(\prm^\star)$}]
\addplot[blue, very thick, mark options={solid,mark size=3}, mark=square] 
      table[x index=0, y index=2, col sep=comma] {./Tikz_Syn/allerton_m1000d50_curv.csv};
      \addlegendentry{CIAG};
\addplot[black, very thick, mark options={solid,mark size=3}, mark=diamond] 
      table[x index=0, y index=2, col sep=comma] {./Tikz_Syn/allerton_m1000d50_sag.csv};
      \addlegendentry{IAG};
\addplot[black!50, very thick, mark options={solid,mark size=3}, mark=o] 
      table[x index=0, y index=2, col sep=comma] {./Tikz_Syn/allerton_m1000d50_igd.csv};
      \addlegendentry{IG};
\addplot[green!80!black, very thick, mark options={solid,mark size=3}, mark=*] 
      table[x index=0, y index=2, col sep=comma] {./Tikz_Syn/allerton_m1000d50_incn.csv};
      \addlegendentry{IN};
\addplot[red, very thick, mark options={solid,mark size=3}, mark=triangle] 
      table[x index=0, y index=2, col sep=comma] {./Tikz_Syn/allerton_m1000d50_bfgs.csv};
      \addlegendentry{IQN};      
\end{semilogyaxis}

\end{tikzpicture}
} }
\else
\resizebox{.98\linewidth}{!}
{ \sf 
\begin{tikzpicture}
\pgfplotsset{ scale only axis,
    width=0.55\textwidth,height=0.3\textwidth,
    grid=both,grid style={line width=.01pt, draw=gray!30},
    legend cell align=left, legend style={legend pos=south east,font=\large, yshift = 0.4cm},
    xlabel={\large Effective Pass ($k/m$)},
    enlarge x limits=0.05,
    enlarge y limits=0.05,
    xmin = 1.5, xmax = 25,
}

\pgfplotsset{every tick label/.append style={font=\large}}

\begin{semilogyaxis}[ylabel shift = -0.1cm, ylabel={\large Opt.~Gap $F(\prm^k) - F(\prm^\star)$}]
\addplot[blue, very thick, mark options={solid,mark size=3}, mark=square] 
      table[x index=0, y index=2, col sep=comma] {./Tikz_Syn/allerton_m1000d50_curv.csv};
      \addlegendentry{CIAG};
\addplot[black, very thick, mark options={solid,mark size=3}, mark=diamond] 
      table[x index=0, y index=2, col sep=comma] {./Tikz_Syn/allerton_m1000d50_sag.csv};
      \addlegendentry{IAG};
\addplot[black!50, very thick, mark options={solid,mark size=3}, mark=o] 
      table[x index=0, y index=2, col sep=comma] {./Tikz_Syn/allerton_m1000d50_igd.csv};
      \addlegendentry{IG};
\addplot[green!80!black, very thick, mark options={solid,mark size=3}, mark=*] 
      table[x index=0, y index=2, col sep=comma] {./Tikz_Syn/allerton_m1000d50_incn.csv};
      \addlegendentry{IN};
\addplot[red, very thick, mark options={solid,mark size=3}, mark=triangle] 
      table[x index=0, y index=2, col sep=comma] {./Tikz_Syn/allerton_m1000d50_bfgs.csv};
      \addlegendentry{IQN};      
\end{semilogyaxis}

\end{tikzpicture}
} 
\fi
\caption{Convergence of the incremental methods for logistic regression problem
of dimension $m=1000$, $d=51$. The $y$-axis denotes the optimality
gap plotted in log-scale and the $x$-axis shows the number of effective passes 
(defined as $k/m$), \ie
number of completed cycles through the $m$ component functions.} \label{fig:small}
\end{figure}
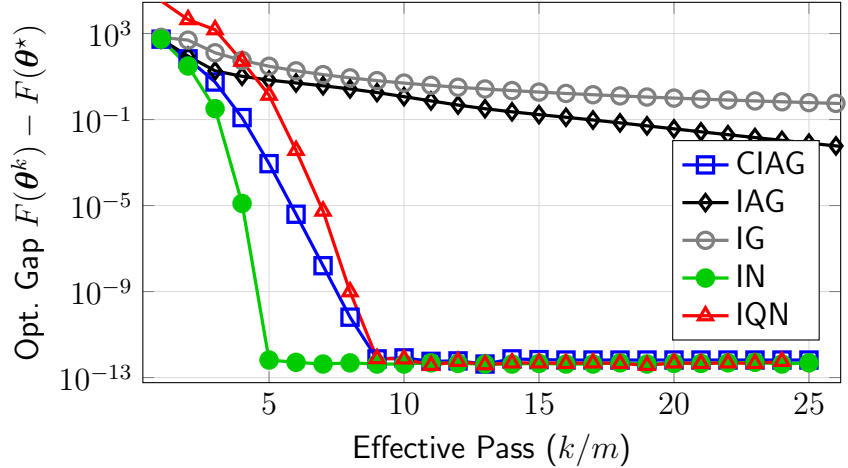

Fig.~\ref{fig:small} 
compares the primal optimality gap\footnote{The optimal objective value
$F(\prm^\star)$ is found by applying the full Newton method on \eqref{eq:opt}
until the latter converges, \ie when $\| \grd F( \prm^k ) \| \leq 10^{-13}$.}, 
$F(\prm^k) - F(\prm^\star)$,
against the iteration number $k$ (notice that $k / m$ is the effective 
number of passes through the component functions) for a 
logistic regression problem instance of dimensions $d=51$ and $m = 1000$. 
Observe that all the incremental methods tested except {\sf IG} converge at least linearly,
and both the {\sf IN} and {\sf IQN} methods appear to converge 
at a superlinear rate, since the slope increases as the iteration number grows.
This corroborates with the claims in \cite{rodomanov2016superlinearly,mokhtari2017iqn}.
We also see that the curvature information aided methods outperform
the others, with the {\sf IN} method being the fastest. 
Even though the {\sf CIAG} method converges only linearly, 
it reaches the same level of optimality ($\approx 10^{-13}$) 
as the superlinearly converging 
{\sf IQN} method using the same number of iterations.

\begin{figure}[t]
\centering
\ifplainver
\resizebox{.7\linewidth}{!}{\resizebox{.98\linewidth}{!}
{ \sf 
\begin{tikzpicture}
\pgfplotsset{ scale only axis,
    width=0.55\textwidth,height=0.3\textwidth,
    grid=both,grid style={line width=.01pt, draw=gray!30},
    legend cell align=left, legend style={legend pos=south east,font=\large, yshift = 0.4cm},
    xlabel={\large Effective Pass ($k/m$)},
    enlarge x limits=0.05,
    enlarge y limits=0.05,
    xmin = 1.5, xmax = 25,
}

\pgfplotsset{every tick label/.append style={font=\large}}

\begin{semilogyaxis}[ylabel shift = -0.1cm, ylabel={\large Opt.~Gap $F(\prm^k) - F(\prm^\star)$}]
\addplot[blue, very thick, mark options={solid,mark size=3}, mark=square, each nth point={2}] 
      table[x index=0, y index=2, col sep=comma] {./Tikz_Syn/allerton_m2000d500_curv.csv};
      \addlegendentry{CIAG};
\addplot[black, very thick, mark options={solid,mark size=3}, mark=diamond, each nth point={2}] 
      table[x index=0, y index=2, col sep=comma] {./Tikz_Syn/allerton_m2000d500_sag.csv};
      \addlegendentry{IAG};
\addplot[black!50, very thick, mark options={solid,mark size=3}, mark=o, each nth point={2}] 
      table[x index=0, y index=2, col sep=comma] {./Tikz_Syn/allerton_m2000d500_igd.csv};
      \addlegendentry{IG};
\addplot[green!80!black, very thick, mark options={solid,mark size=3}, mark=*, each nth point={2}] 
      table[x index=0, y index=2, col sep=comma] {./Tikz_Syn/allerton_m2000d500_incn.csv};
      \addlegendentry{IN};
\addplot[red, very thick, mark options={solid,mark size=3}, mark=triangle, each nth point={2}] 
      table[x index=0, y index=2, col sep=comma] {./Tikz_Syn/allerton_m2000d500_bfgs.csv};
      \addlegendentry{IQN};      
\end{semilogyaxis}

\end{tikzpicture}
} }
\else
\resizebox{.98\linewidth}{!}
{ \sf 
\begin{tikzpicture}
\pgfplotsset{ scale only axis,
    width=0.55\textwidth,height=0.3\textwidth,
    grid=both,grid style={line width=.01pt, draw=gray!30},
    legend cell align=left, legend style={legend pos=south east,font=\large, yshift = 0.4cm},
    xlabel={\large Effective Pass ($k/m$)},
    enlarge x limits=0.05,
    enlarge y limits=0.05,
    xmin = 1.5, xmax = 25,
}

\pgfplotsset{every tick label/.append style={font=\large}}

\begin{semilogyaxis}[ylabel shift = -0.1cm, ylabel={\large Opt.~Gap $F(\prm^k) - F(\prm^\star)$}]
\addplot[blue, very thick, mark options={solid,mark size=3}, mark=square, each nth point={2}] 
      table[x index=0, y index=2, col sep=comma] {./Tikz_Syn/allerton_m2000d500_curv.csv};
      \addlegendentry{CIAG};
\addplot[black, very thick, mark options={solid,mark size=3}, mark=diamond, each nth point={2}] 
      table[x index=0, y index=2, col sep=comma] {./Tikz_Syn/allerton_m2000d500_sag.csv};
      \addlegendentry{IAG};
\addplot[black!50, very thick, mark options={solid,mark size=3}, mark=o, each nth point={2}] 
      table[x index=0, y index=2, col sep=comma] {./Tikz_Syn/allerton_m2000d500_igd.csv};
      \addlegendentry{IG};
\addplot[green!80!black, very thick, mark options={solid,mark size=3}, mark=*, each nth point={2}] 
      table[x index=0, y index=2, col sep=comma] {./Tikz_Syn/allerton_m2000d500_incn.csv};
      \addlegendentry{IN};
\addplot[red, very thick, mark options={solid,mark size=3}, mark=triangle, each nth point={2}] 
      table[x index=0, y index=2, col sep=comma] {./Tikz_Syn/allerton_m2000d500_bfgs.csv};
      \addlegendentry{IQN};      
\end{semilogyaxis}

\end{tikzpicture}
} 
\fi
\caption{Convergence of the incremental methods for logistic regression problem
of dimension $m=2000$, $d=501$. Note that the {\sf IN} method is not 
converging in this example.} \label{fig:large}
\end{figure}
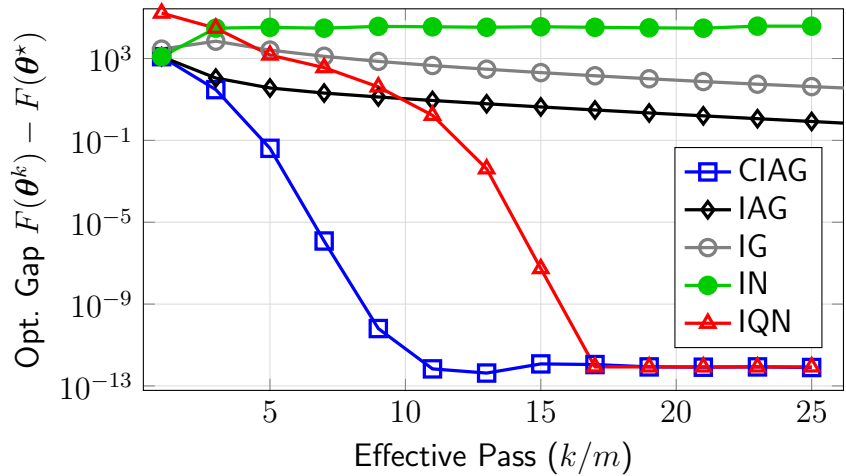

Fig.~\ref{fig:large} considers an instance of a similar problem 
with larger dimensions of $d=501$ and $m=2000$. 
We observe that the {\sf IN} method is \emph{not} converging in this problem
instance. In fact, in our experiments, the {\sf IN} method 
using a step size of $\gamma = 1$ is often not converging on 
random problem instances with large $d$ and $m$. 
On the other hand, both the {\sf CIAG} and {\sf IQN} methods  
appear to have significantly better numerical stability. 
Moreover, the {\sf CIAG} method has the fastest
 convergence
among the incremental methods tested, even though its theoretical convergence 
rate is at most linear, as opposed to the superlinear convergence rate 
of the {\sf IQN} method. 

Our second set of experiments focuses on training an SVM from real training data. 
The incremental methods are implemented in \texttt{c++}. We use the 
implementations of {\sf IAG} and {\sf IN} from \cite{rodomanov2016superlinearly} 
with the {\sf IN} method set
to compute the inverse of the aggregated Hessian in an exact fashion.
The training datasets used are \texttt{mushrooms} ($m=8124$, $d=112$), 
\texttt{w8a} ($m=49749$, $d=300$) and 
\texttt{alpha} ($m=500000$, $d=500$)
from \texttt{libsvm} \cite{CC01a}. 
Notice that as both $m$ and $d$ are large in these cases,
the {\sf IQN} method run out of memory while initializing its working variables. 
We use a minibatch setting such that $5$ samples are selected
for every iteration of the incremental methods. 
Lastly, we choose the step sizes for {\sf CIAG} as $\gamma = 0.001 / L$ 
for \texttt{mushrooms}, \texttt{w8a} and $\gamma = 10^{-5} / L$ for \texttt{alpha};
for {\sf IAG} as $\gamma = 0.1 / (mL)$ for \texttt{mushrooms}, 
$\gamma = 0.05 / (mL)$ for \texttt{w8a} and $\gamma = 10^{-5} / (mL)$ 
for \texttt{alpha}. 

In Table~\ref{tab:real}, we compare the 
performance in terms of the running time and number of required iterations
to reach convergence, \ie when $\| \grd F(\prm^k ) \| \leq 10^{-10}$. 
We observe that the {\sf IN} method has the fastest convergence rate, \ie
requiring the least number of iterations (or effective passes), for all
the problem instances. 
However, except for the low-dimension problem case (e.g., \texttt{mushrooms})
where $d \approx 100$,
the {\sf IN} method requires a much longer running time
than the proposed {\sf CIAG} method to reach convergence. 
This is due to the added complexity required for computing the $d \times d$ 
Hessian inverse. 
These experimental findings corroborate our claim that the {\sf CIAG} method 
achieves a desirable tradeoff between complexity and convergence speed. 

\begin{table}[t]
\begin{center}
\renewcommand\baselinestretch{1.2}\selectfont
\begin{tabular}{c||c||c||c} 
\hline
Dataset & {\sf CIAG} method & {\sf IN} method \cite{rodomanov2016superlinearly} & {\sf IAG} method \cite{gurbuzbalaban2017convergence} \\
\hline
\hline 
\multirow{2}{*}{\texttt{mushrooms}} & $43.5$ eff.~pass & \bfseries $\bf 4.8$ eff.~pass & $1920$ eff.~pass \\
& $2.256$ sec. & \bfseries $\bf 1.002$ sec. & $6.848$ sec. \\
\hline
\multirow{2}{*}{\texttt{w8a}} & \bfseries $\bf 7.2$ eff.~pass & $5.3$ eff.~pass & $\geq 10^3$ eff.~pass \\
& \bfseries $\bf 16.38$ sec. & $64.62$ sec. & $\geq 75.83$ sec.$^\dagger$ \\
\hline
\multirow{2}{*}{\texttt{alpha}} & \bfseries $\bf 7.6$ eff.~pass & $2.3$ eff.~pass & 
$\geq 10^3$ eff.~pass \\[-.05cm]
& \bfseries $\bf 463.08$ sec. & 1130.7 sec. & $\geq 1353.8$ sec.$^\dagger$ \\
\hline
\end{tabular}\vspace{.2cm}
\end{center}
\caption{Performance of the incremental methods on different training datasets. 
We show the number of effective passes (defined as $k/m$) 
and the corresponding running time required to reach convergence 
such that $\| \grd F(\prm^k) \| \leq 10^{-10}$. ($^\dagger$the {\sf IAG} method
only converges to a solution with $\| \grd F(\prm^k) \| \approx 10^{-6}$ [for \texttt{w8a}]
and $\| \grd F(\prm^k) \| \approx 10^{-4}$ [for \texttt{alpha}] after
$1000$ effective passes).} \label{tab:real} \vspace{-0.5cm}
\end{table}

\section{Conclusions}
In this paper, we have proposed a new incremental gradient-type method that
uses curvature information to accelerate convergence. 
We show that the proposed {\sf CIAG} method converges linearly
with a rate comparable to that of the {\sf FG} method, while requiring 
a single \emph{incremental} step only.
Numerical results are presented to support our theoretical claims. 
Future work includes incorporating the Nesterov's acceleration into
the {\sf CIAG} update, analyzing the convergence speed for {\sf CIAG}
with randomized
component selection scheme, relaxing the strong convexity assumption, etc.
The recently developed double {\sf IAG} technique developed in 
\cite{mokhtari2016surpassing} may also be adopted into the {\sf CIAG}
method for further acceleration. 

\section{Acknowledgement}
This work is supported by 
NSF CCF-BSF 1714672.

\bibliographystyle{IEEEtran}
\bibliography{cigd_ref}

\ifplainver
\newpage
\section{Proof of Theorem~\ref{thm:scvx}} \label{sec:pf}
Our idea is to analyze the {\sf CIAG}  method as a \emph{noisy} 
gradient descent method. In particular, let us define
\beq 
\tilde{\bm g}^k \eqdef \sum_{i=1}^m \Big( 
\grd f_i ( \prm^{\tau_i^k} ) + \grd^2 f_i( \prm^{\tau_i^k} ) ( \prm^k - \prm^{\tau_i^k} ) \Big) 
\eeq
as the gradient surrogate employed in {\sf CIAG}. 
Define ${\bm e}^k \eqdef \tilde{\bm g}^k - \grd F(\prm^k)$ as the error between
the exact gradient and the gradient surrogate used by {\sf CIAG}.
We define the following as our optimality measure:
\beq
\dst{k} \eqdef \| \prm^k - \prm^\star \|^2 \eqs.
\eeq
Recall that $\tilde{\bm g}^k = \grd F(\prm^k) + {\bm e}^k$ and let us define:
\beq
E(k) \eqdef \gamma^2 \| {\bm e}^k \|^2 - 2 \gamma \langle \prm^k - \prm^\star - \gamma \grd F(\prm^k), {\bm e}^k \rangle \eqs.
\eeq 
We observe the following chain\footnote{This chain of analysis follows from \cite[Section 3.3]{gurbuzbalaban2017convergence} and is repeated here 
merely for
the sake of self-containedness.}:
\beq \label{eq:gd}
\begin{split}
\dst{k+1} & = \dst{k} - 2 \gamma \langle \grd F(\prm^k), \prm^k - \prm^\star \rangle \\
& \hspace{2.75cm} + \gamma^2 \| \grd F(\prm^k) \|^2 + E(k) \\
& \hspace{-1cm} \leq \dst{k} - 2 \gamma \Big( \frac{\mu L}{\mu+L} \dst{k} + \frac{1}{\mu + L} \| \grd F(\prm^k) \|^2 \Big) \\
& \hspace{2.75cm} + \gamma^2 \| \grd F(\prm^k) \|^2 + E(k) \\
& \hspace{-1cm} = \Big( 1 - 2\gamma  \frac{ \mu L }{\mu + L}\Big) \dst{k} + E(k) \\
& \hspace{2cm} + \Big( \gamma^2 - \frac{2 \gamma}{\mu+L} \Big) \| \grd F(\prm^k) \|^2 \\
& \hspace{-1cm} \leq \Big( 1 - 2\gamma  \frac{ \mu L }{\mu + L}\Big) \dst{k} + E(k) \eqs,
\end{split}
\eeq
where the first inequality is due to Assumption~\ref{ass:cts} and \ref{ass:strcvx}
and the last inequality is due to the choice of step size that $\gamma < 2 / (\mu + L)$ [cf.~\eqref{eq:step}]. 

The next step is to bound the error term $E(k)$. We observe that:
\beq \label{eq:ek0} \begin{split}
|E(k)| & \leq \gamma^2 \| {\bm e}^k \|^2 + 2 \gamma \| {\bm e}^k \| \| \prm^k - \gamma \grd F(\prm^k) - \prm^\star \| \\
& \leq \gamma^2 \| {\bm e}^k \|^2 + 2 \gamma \sqrt{\dst{k}} \| {\bm e}^k \| \eqs,
\end{split}
\eeq
where the last inequality is due to the fact that 
the term $\prm^k - \gamma \grd F(\prm^k)$ is equivalent to applying
an exact gradient descent step to $\prm^k$. It follows that the conclusion in \eqref{eq:gd} 
holds with $E(k) = 0$, therefore the difference between the term and
$\prm^\star$ cannot be greater than $\dst{k}$ due to the choice of our step size.
Moreover, we observe that
\beq \label{eq:ek}
\begin{split}
\| {\bm e}^k \| & \leq \sum_{i=1}^m L_{H,i} \| \prm^{\tau_i^k} - \prm^k \|^2 \eqs,
\end{split}
\eeq
where the inequality is due to Assumption~\ref{ass:cts_h}. 
Importantly, we observe that the error's norm is bounded by a \emph{squared}
norm of the difference $\prm^{\tau_i^k} - \prm^k$. 
Proceeding from \eqref{eq:ek}, we can bound $|E(k)|$ as:
\beq \label{eq:ek2}
\begin{split}
& |E(k)| \\
& \leq \gamma^6 8K^4 L_H^2 \Big( \max_{ k' \in [k-2K, k] } \big(L^4 \dst{k'}^2 + 256 L_H^4  \dst{k'}^4 \big) \Big) \\
& ~~+ \gamma^3 4K^2 L_H \Big( \max_{k' \in [k-2K,k]} \big(L^2 \dst{k'}^{\frac{3}{2}} + 16 L_H^2 \dst{k'}^{\frac{5}{2}} \big) \Big) \eqs,
\end{split}
\eeq
the derivation of the above can be found in Section~\ref{sec:ek}. 

We can now conclude the proof by substituting $\dst{k} = R(k)$ in Lemma~\ref{lem:2},
where the proof of the latter can be found in Section~\ref{sec:lem}. 
Our 
sequence $\{ \dst{k} \}_k$ is non-negative and it satisfies \eqref{eq:ineq2} with:
\beq
\begin{split}
p  & = 1 - 2 \gamma \mu L / (\mu + L) \eqs,\\
q_1 & = \gamma^6 \cdot 8 K^4 L_H^2 L^4  \eqs,\\
q_2 & = \gamma^6 \cdot 2048 K^4 L_H^6 \eqs,\\
q_3 & = \gamma^3 \cdot 4 K^2 L_H L^2 \eqs,\\
q_4 & = \gamma^3 \cdot 64 K^2 L_H^3 \eqs,
\end{split}
\eeq 
and $\eta_1 = 2$, $\eta_2 = 4$, $\eta_3 = 3/2$, $\eta_4 = 5/2$, $M = 2K+1$. 
To satisfy \eqref{eq:cond2}, we require the following on the step size $\gamma$:
\beq
\begin{split}
& ~\textstyle p + \sum_{j=1}^4 q_j V(0)^{\eta_j-1} < 1 \\[.1cm]
\Longleftrightarrow & ~\textstyle \sum_{j=1}^4 q_j V(0)^{\eta_j - 1} < 2 \gamma \mu L / (\mu+L) \\[.1cm]
\Longleftrightarrow & ~\gamma^5 \cdot 8K^4 L_H^2 \Big( L^4 V(0) + 256 L_H^4 V(0)^3 \Big) \\
& ~+ \gamma^2 \cdot 4K^2 L_H \Big( L^2 V(0)^{1/2} + 16 L_H^2 V(0)^{3/2} \Big) \\
& ~< 2 \mu L / (\mu + L) \eqs,
\end{split}
\eeq
which can be satisfied by having:
\beq \notag
\gamma^5 \cdot 8K^4 L_H^2 \Big( L^4 V(0) + 256 L_H^4 V(0)^3 \Big) < \frac{ \mu L } { \mu + L } ~~\text{and}
\eeq
\beq \notag
\gamma^2 \cdot 4K^2 L_H \Big( L^2 V(0)^{1/2} + 16 L_H^2 V(0)^{3/2} \Big) < \frac{ \mu L } { \mu + L } \eqs.
\eeq
The above can be satisfied by the step size choice in \eqref{eq:step}. 
As such, the conclusions in the lemma hold 
and the conclusions of Theorem~\ref{thm:scvx}
follow.

\subsection{Derivation of Eq.~\eqref{eq:ek2}} \label{sec:ek}
We observe that:
\beq \label{eq:tmp_eq}
\begin{split}
\| {\bm e}^k \| & \leq \sum_{i=1}^m L_{H,i} \| \prm^{\tau_i^k} - \prm^k \|^2 \\
& \leq \sum_{i=1}^m L_{H,i} 
\underbrace{(k - \tau_i^k)}_{\leq K} \sum_{j=\tau_i^k}^{k-1} \| \prm^{j+1} - \prm^j \|^2 \\
& \leq K L_H \sum_{j= (k-K)_+}^{k-1}  \| \prm^{j+1} - \prm^j \|^2 \\
& \leq K L_H \gamma^2 \sum_{j=(k-K)_+}^{k-1} \| {\bm e}^j + \grd F(\prm^j) \|^2 \\
& \leq 2 \gamma^2 K L_H \sum_{j=(k-K)_+}^{k-1} \big( \| {\bm e}^j \|^2 + \| \grd F(\prm^j) \|^2 \big)
\end{split}
\eeq
We have
\beq
\| \grd F(\prm^j) \|^2 = \| \grd F(\prm^j) - \grd F(\prm^\star) \|^2 \leq L^2 \dst{j} \eqs,
\eeq
and
\beq \begin{split}
\| {\bm e}^j \| & \textstyle \leq \sum_{i=1}^m L_{H,i} \| \prm^j - \prm^{\tau_i^j} \|^2 \\
& \textstyle \leq 2 \sum_{i=1}^m L_{H,i} 
\cdot \big( \dst{j} + \dst{\tau_i^j} \big) \\
& \textstyle \leq 4 L_H \max_{ \ell \in \{ \tau_i^j \}_{i=1}^m \cup \{j\} } \dst{\ell}
\end{split}
\eeq
Plugging these back into \eqref{eq:tmp_eq} gives:
\beq \begin{split}
& \| {\bm e}^k \| \\
& \leq 2 \gamma^2 K L_H \sum_{j=(k-K)_+}^{k-1} \Big( L^2 \dst{j} + 
\big( 4 L_H \max_{ \ell \in \{ \tau_i^j \}_{i=1}^m \cup \{j\} } \dst{\ell} \big)^2 \Big) \\
& \textstyle \leq 2 \gamma^2 K^2 L_H \Big( L^2 \max_{ (k-K)_+ \leq \ell \leq k-1 } \dst{\ell} \\
& \textstyle \hspace{3cm} + 16 L_H^2 
\max_{ (k-2K)_+ \leq \ell \leq k-1 } \dst{\ell}^2 \Big) \eqs,
\end{split}
\eeq
where we have used the fact that $\tau_i^{k-K} \geq k - 2K$
in the last inequality.
Consequently, we can upper bound $E(k)$ as 
\beq \notag
\begin{split}
|E(k)| & \leq 4 \gamma^6 K^4 L_H^2 \Big( L^2 \max_{ (k-K)_+ \leq \ell \leq k-1 } \dst{\ell} 
\\
& \hspace{2.5cm} + 16 L_H^2 \max_{ (k-2K)_+ \leq \ell \leq k-1 } \dst{\ell}^2 \Big)^2  \\
& ~~~~+  4 \gamma^3 K^2 L_H \cdot \sqrt{\dst{k}} \cdot \Big( L^2 \max_{ (k-K)_+ \leq \ell \leq k-1 } \dst{\ell} \\
& \hspace{2.5cm}
+ 16 L_H^2 \max_{ (k-2K)_+ \leq \ell \leq k-1 } \dst{\ell}^2 \Big) \eqs,
\end{split}
\eeq
which can be further bounded as \eqref{eq:ek2}.

\subsection{Proof of Lemma~\ref{lem:2}}\label{sec:lem}
The proof of Lemma~\ref{lem:2} is divided into two parts. 
We first show that under \eqref{eq:cond2}, the sequence 
$R(k)$ converges linearly as in \eqref{eq:lem1}; then we show
that the rate of convergence can be improved to $p$
as in \eqref{eq:lem2}. 

The first part of the proof is achieved using induction on all $\ell \geq 1$ with the following 
statement: 
\beq \label{eq:ind}
R(k) \leq \delta^\ell \cdot R(0),~\forall~k=(\ell-1)M + 1,..., \ell M \eqs.
\eeq 
The base case when $\ell=1$ can be straightforwardly established:
\beq
\begin{split}
& \textstyle R(1) \leq p R(0) + \sum_{j=1}^J q_j R(0)^{\eta_j} \leq \delta R(0) \eqs, \\
& \textstyle R(2) \leq p R(1) + \sum_{j=1}^J q_j R(0)^{\eta_j} \leq \delta R (0) \eqs, \\
& \vdots \\
& \textstyle R(M) \leq p R(M-1) \sum_{j=1}^J q_j R(0)^{\eta_j} \leq \delta R (0) \eqs. 
\end{split}
\eeq
Now suppose that the statement \eqref{eq:ind} is true up to $\ell=c$, for $\ell=c+1$,
we have:
\beq
\begin{split}
R( cM+ 1) & \leq p R( cM ) + \sum_{j=1}^J q_j \max_{ k' \in [ (c-1)M + 1, cM ] } R(k')^{\eta_j} \\
& \leq p \big( \delta^c R(0) \big) + \sum_{j=1}^J q_j  \big( \delta^c R(0) \big)^{\eta_j} \\
& \leq \delta^c \cdot \Big( pR(0) + \sum_{j=1}^J q_j  R(0)^{\eta_j} \Big) \leq \delta^{c+1} R(0) \eqs.
\end{split}
\eeq 
Similar statement also holds for $R(k)$ with $k=cM+2,...,(c+1)M$. We thus conclude that:
\beq
R(k) \leq \delta^{ \lceil k / M \rceil } \cdot R(0),~\forall~ k \geq 0 \eqs,
\eeq
which proves \eqref{eq:lem1}. 

The second part of the proof establishes the linear rate of convergence of $p$. We
observe that
\beq \label{eq:sec_part}
\frac{R(k+1)}{R(k)} \leq  p + \frac{ \sum_{j=1}^J q_j \max_{ k' \in [(k-M+1)_+, k] } R(k')^{\eta_j} }{ R(k) } \eqs.
\eeq
For any $k' \in [k-M+1,k]$ and any $\eta > 1$, we have:
\beq \begin{split}
& \frac{ R(k')^{\eta} }{ R(k) } \leq \frac{ R(k') }{R(k) } R(0)^{\eta-1} \delta^{ (\lceil \frac{k'}{M} \rceil)(\eta-1) }
\end{split}
\eeq
As $\eta > 1$, we observe that $\delta^{ (\lceil \frac{k'}{M} \rceil)(\eta-1) } \rightarrow 0$ 
when $k \rightarrow \infty$.
We have two cases to be analyzed --- the first case is 
\beq \label{eq:case1}
\lim_{k \rightarrow \infty} \frac{ R(k') }{R(k) } = \infty \Longleftrightarrow
\lim_{k \rightarrow \infty} \frac{R(k) }{ R(k') } = 0 \eqs,
\eeq
Assume that $\lim_{k \rightarrow \infty} {R(k) } / { R(k-1) }$ is well defined,
for $k' \neq k$, we observe that:
\beq \label{eq:case1e} \begin{split}
& \lim_{k \rightarrow \infty} \frac{R(k) }{R(k-1)} \frac{R(k-1)}{R(k-2)} \cdots \frac{R(k'+1)}{ R(k') } = 0 \\ 
\Longrightarrow~ & \Big( \lim_{k \rightarrow \infty} \frac{R(k+1) }{R(k)} \Big)^{k-k'} = 0  \\
\Longrightarrow~ & \lim_{k \rightarrow \infty} \frac{R(k+1) }{R(k)} = 0 \eqs,
\end{split}
\eeq
where we have used the property for the limit of products.
On the other hand, when $k'=k$ it leads to a contradiction to \eqref{eq:case1} 
since $\lim_{k \rightarrow \infty} R(k)/ R(k) = 1$.

The second case is that
\beq \label{eq:case2}
\lim_{k \rightarrow \infty} { R(k')^{\eta} } / { R(k) } = 0 \eqs.
\eeq
Notice that if \eqref{eq:case1} happens for some $k' \in [k-M+1,k]$, then the derivation in 
\eqref{eq:case1e} implies that $\lim_{k \rightarrow \infty} R(k+1)/R(k) = 0 \leq p$.
Alternatively, if \eqref{eq:case2} happens for \emph{all} $k' \in [k-M+1,k]$, 
then \eqref{eq:sec_part} implies that
\beq
\lim_{k \rightarrow \infty} R(k+1) / R(k) \leq p \eqs,
\eeq
this concludes our proof. 
\else
\fi


\end{document}